\documentclass[11pt]{article}
\usepackage{booktabs}

\usepackage[preprint]{acl}

\usepackage{times}
\usepackage{latexsym}

\usepackage[T1]{fontenc}

\usepackage[utf8]{inputenc}

\usepackage{microtype}

\usepackage{inconsolata}

\usepackage{graphicx}

\usepackage{xcolor}

\usepackage{amssymb}
\usepackage{amsmath}

\usepackage{tikz}
\usetikzlibrary{arrows.meta,positioning}

\usepackage{pgfplots}
\usepgfplotslibrary{groupplots}
\pgfplotsset{compat=1.17}

\usepackage{multirow}

\usepackage{algorithm}
\usepackage{algpseudocode}

\newcommand{\joingr}{\textsc{JoinGR}\xspace}
\usepackage{xspace}
 
\title{\joingr: Learning to Traverse Join Graphs for \\
       Table Retrieval}

\author{
\textbf{Sandipan De}\textsuperscript{1}
\quad
\textbf{Abhijit Chakraborty}\textsuperscript{1}
\quad
\textbf{Sambaran Bandyopadhyay}\textsuperscript{2}
\quad
\textbf{Vivek Gupta}\textsuperscript{1}
\\[2pt]
Arizona State University\textsuperscript{1}
\quad
Adobe Research, India\textsuperscript{2}
\\[4pt]
\texttt{sandipan@asu.edu, achakr40@asu.edu, samb.bandyo@gmail.com, vgupt140@asu.edu}
}

\begin{document}
\maketitle

\begin{abstract}
Retrieving the right tables is a prerequisite for Text-to-SQL over realistic databases. Dense table retrievers rank schema elements independently, but this ignores a key source of evidence: some required tables are not mentioned in the question and become identifiable only through their join relationships to already relevant tables. We introduce \joingr, a join-aware table retrieval method that treats the database join graph as the retrieval space. Columns are represented as graph nodes, while intra-table and foreign-key relationships are represented as typed edges. Given a question, \joingr selects semantically similar anchor tables, traverses join edges with a query-conditioned scorer, and aggregates the resulting edge deposits into table scores. The scorer is a lightweight MLP on top of frozen query, node, and edge embeddings, trained with a pairwise margin loss over gold tables. On BIRD and Spider datasets, \joingr is competitive with the strongest retrieval baselines. On BEAVER, a challenging enterprise benchmark with multi-hop table requirements, \joingr substantially improves recall over dense retrieval and re-ranking baselines. Cross-domain experiments show that the learned scorer transfers across benchmarks, indicating that the method captures reusable join-graph traversal behavior.

\end{abstract}

\section{Introduction}
\label{sec:intro}
Modern text-to-SQL systems increasingly rely on large language models to perform semantic parsing \cite{li2023bird, pourreza2023din}. However, their success still depends on an upstream retrieval step: the model must have access to the schemas of the relevant table(s). 
In realistic databases, the full schema is often too large to place in context, and irrelevant schema elements can induce spurious joins or hallucinated SQL \cite{maamari2024the}. 
Conversely, if even one required table is missing from the retrieved context, the correct query may be impossible to generate. Table retrieval therefore narrows the schema before downstream schema linking and generation.

The dominant approach is dense retrieval. A question and schema elements are embedded into a shared vector space, and tables are ranked by semantic similarity to the question. This works well when schemas are compact, table and column names are descriptive, and when questions refer directly to the required tables. However, independent retrieval struggles when queries require multi-hop joins, or when table names are abbreviated, overlapping, or meaningful only in relation to other tables. A bag-of-columns retriever has no mechanism for expressing that a table becomes relevant only after another table has been selected.

We argue that this \emph{structural blindness}, rather than embedding quality, is a key constraint. 
To make schema relationships first-class, we posit \emph{Join Graph}: a graph representation in which columns are nodes and intra-table and foreign-key relationships are typed edges.

\begin{figure*}[t]  \includegraphics[width=\textwidth]{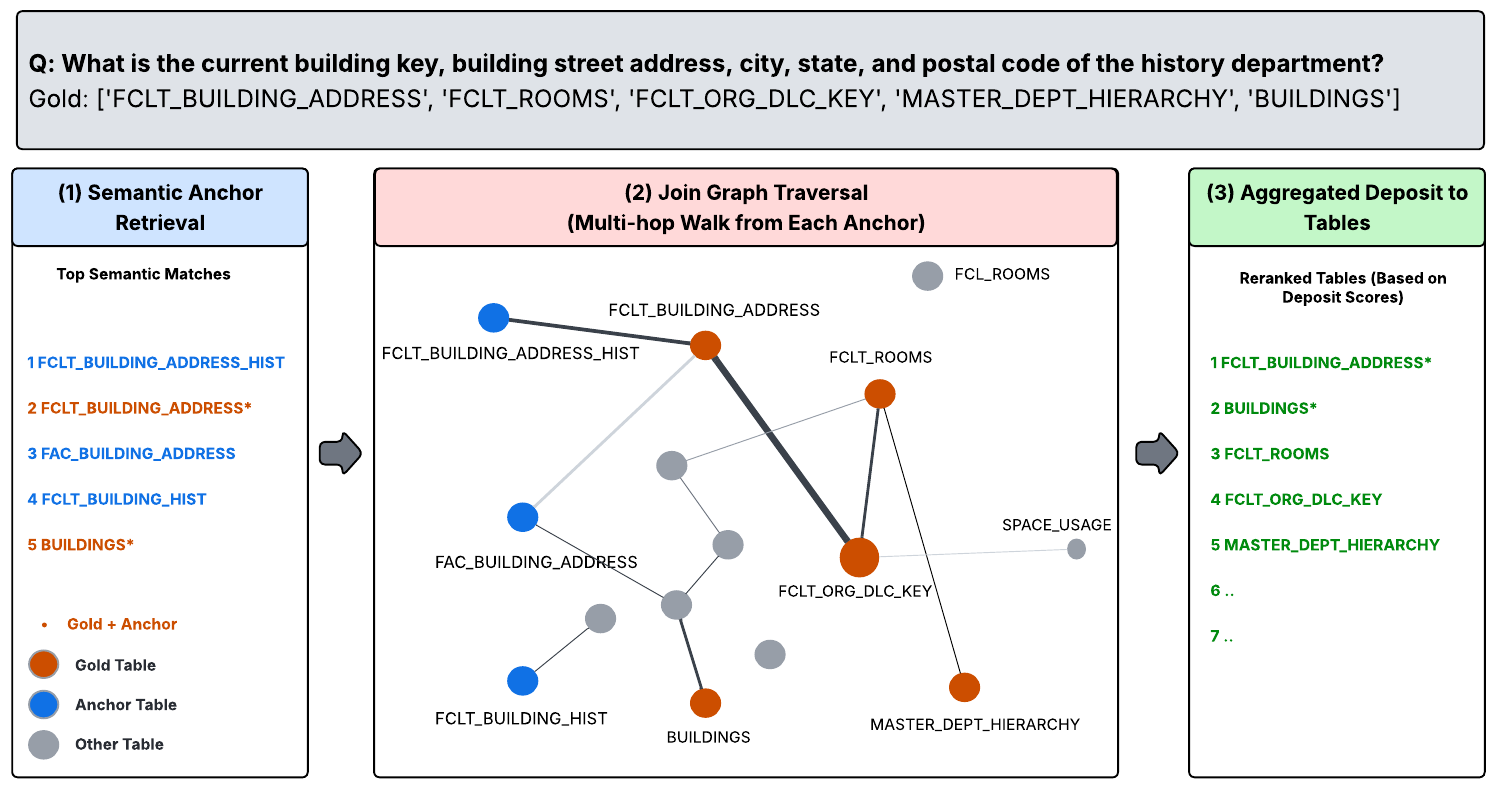}  \caption{Overview of \joingr{} table retrieval for the example question. (1) Semantic retrieval selects the top-5 anchor tables based on dense similarity between the query and table/column representations. (2) Starting from these anchors, \joingr performs query-conditioned multi-hop traversal over the join graph; edge thickness indicates traversal probability and node size reflects aggregated deposit received during propagation. Gold tables are highlighted in orange. (3) Deposits accumulated from all traversal paths are aggregated to produce final table scores.}    \label{fig:method}\end{figure*}

Structure alone is insufficient as it specifies table connections but not their relevance to specific questions. Different questions from the same anchor table may need different join edges. \joingr combines semantic matching to select anchor tables and a query-conditioned scorer to determine relevance in the join graph. The scorer assigns weights to join edges, directing relevance flow along promising paths. Final table scores aggregate deposits from these query-specific walks. Figure~\ref{fig:method} details the three-stage flow.

We evaluate \joingr on Spider \cite{yu2018spider}, BIRD \cite{li2023bird}, and BEAVER \cite{stonebraker2024beaver}, spanning various retrieval difficulties. Spider offers a standard cross-domain setting where schemas are compact, and dense retrieval is strong. BIRD features larger databases, but many queries need only a few tables. BEAVER is a challenging enterprise benchmark with cryptic table names and multi-hop table needs. This range lets us determine when structure-aware retrieval is needed: \joingr should be competitive when semantic retrieval suffices and improve recall when relational structure is crucial.

Our results support this view. On Spider and BIRD, \joingr is competitive with strong retrieval baselines in regimes where dense embedding similarity already performs well. On BEAVER, where independent retrieval and re-ranking methods struggle, \joingr substantially improves recall and perfect recall. Moreover, scorers trained on one benchmark transfer to another without retraining, suggesting that the model learns to traverses join graphs rather than memorizing dataset-specific table names.

Our contributions are:
\begin{itemize}
    \item We formulate table retrieval as query conditioned traversal over a database join graph, enabling retrieval of tables that are structurally implied rather than directly mentioned.
    \item We propose \joingr{}, an anchor-walk-aggregate method that combines semantic retrieval with learned propagation over join relationships.
    \item We evaluate on Spider, BIRD, and BEAVER (DW), showing that \joingr{} is competitive when dense retrieval is strong and substantially improves recall and perfect recall on multi-hop enterprise schemas.
    \item We show that the learned scorer transfers across benchmarks, supporting its use as a reusable traversal policy for new schemas with limited supervision.
\end{itemize}

\section{Problem Formulation}
\label{sec:problem}


Let $\mathcal{D}=\{D_1,\ldots,D_n\}$ denote a dataset of relational databases.
For a database $D\in\mathcal{D}$, let
$\mathcal{T}(D)=\{t_1,\ldots,t_m\}$ denote its set of tables and
$\mathcal{C}(D)=\bigcup_i \mathcal{C}(t_i)$ denote the set of all columns across those tables. A natural-language question $q$ associated with database $D$ has a gold set
$\mathcal{T}^g(q)\subseteq\mathcal{T}(D)$ of tables required to construct the correct SQL query, and (when annotated) a gold set $\mathcal{K}^g(q)$
of join-key pairs $(c,c^\prime)$, where
$c\in\mathcal{C}(t_i)$ and
$c^\prime\in\mathcal{C}(t_j)$ for $i\neq j$.

The table retrieval task requires: given $q$, produce a ranked list of tables that contains all of $\mathcal{T}^g(q)$ at as small a rank as possible.

The above formulation treats table retrieval as a ranking problem. 
However, ranking tables independently ignores the fact that some tables necessary to answer a query may not be immediately obvious from the query utterance; instead, they may become relevant only through their structural relationship to already identified tables. We therefore utilize the database relational structure as the retrieval space for \joingr.

\section{\joingr}
\label{sec:method}
\joingr retrieves tables by first constructing a structured search space from the schema and then performing learned traversal over that space. We describe the join-graph representation, followed by the anchor--walk--aggregate retrieval procedure and training objective.

\subsection{Join Graph Construction}
\label{sec:graph}
\begin{figure}[t]
  \includegraphics[width=\columnwidth]{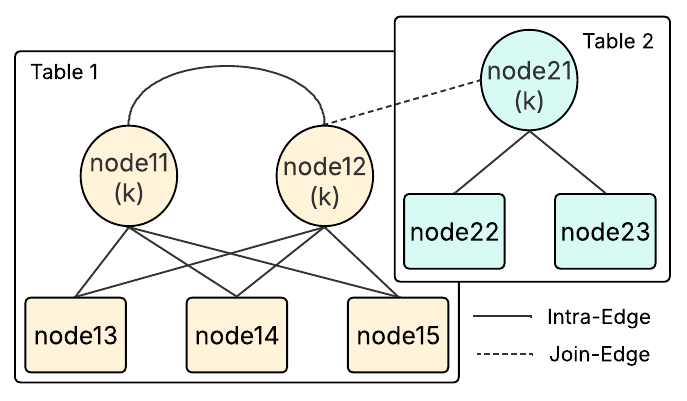}
  \caption{Example of the join graph constructed from a relational schema. 
  Circular nodes denote key columns, rectangular nodes denote non-key columns, solid edges represent intra-table connectivity, and dashed edges represent inter-table join relationships.}
    \label{fig:navigation_graph}
\end{figure}

We construct a single undirected graph
$G_{\mathcal{D}}=(V,E)$ over all databases in the dataset.
Each column $c\in\mathcal{C}(D)$ for some $D\in\mathcal{D}$ is represented as a node $v_c\in V$ with a node descriptor $d_v$ containing
(i)~the canonical name (\texttt{db.table.column}),
(ii)~a short LLM-generated semantic name and \emph{purpose},
(iii)~the data type, and
(iv)~a key flag \texttt{is\_key} indicating whether $c$ is either a primary or foreign key of the table.
The relationship between two columns is represented as an edge $e=(v_c,v_{c'})\in E$ with an edge descriptor $d_e$ containing
(i)~the endpoint node identifiers and names,
(ii)~the edge type, and
(iii)~a short LLM-generated \emph{edge purpose} describing why the two columns are related.
Edges are typed as either \textit{intra} or \textit{join}.
An \textit{intra} edge connects columns belonging to the same table, specifically between a key column and a non-key column or between two key columns. 
A \textit{join} edge connects key columns from different tables that participate in a foreign-key relationship.
Figure~\ref{fig:navigation_graph} illustrates the resulting connectivity patterns induced by \textit{intra} and \textit{join} edges.

\subsection{\texorpdfstring{Anchor $\rightarrow$ Walk $\rightarrow$ Aggregate}{Anchor -> Walk -> Aggregate}}

While the join graph defines the space for possible connections, a query usually requires only a small part of the structure. \joingr retrieves this query-specific subgraph through a three stage process. First, it selects a small set of semantically relevant \emph{anchor tables} using similarity between query and node embeddings. Second, it \emph{walks} from these anchors along the join edges using a \emph{query-conditioned} scorer to decide which paths are most relevant, and third, it aggregates the evidence accumulated on the traversed edges into final table scores.


\subsubsection{Encoding}
\joingr operates in a shared embedding space for queries, nodes, and edges.  Let $\varphi(\cdot)$ be a frozen encoder. For query $q$, we compute $\mathbf{q}=\varphi(q)\in\mathbb{R}^d$. 
Offline, for each node $v$ with descriptor $d_v$, we compute
$\mathbf{v}=\varphi(d_v)\in\mathbb{R}^d$. 
Similarly, for each edge $e$ with descriptor $d_e$, we compute
$\mathbf{e}=\varphi(d_e)\in\mathbb{R}^d$.
All vectors are $\ell_2$-normalized so that dot products correspond to cosine similarities.


\subsubsection{Anchor Selection}

The walk begins from tables that are semantically close to the query.
We first retrieve the nearest nodes from the query embedding and map each retrieved node to its parent table.
Each candidate table is assigned a semantic table score
\[
s(t \mid q) = \max_{v \in V(t)} \langle \mathbf{q}, \mathbf{v} \rangle .
\label{eq:table-score}
\]
We use $s(t \mid q)$ to rank candidate tables and initialize anchor deposits.\footnote{
Max similarity avoids diluting a highly indicative column among many unrelated columns in wide tables.
}
For later use in the graph walk, we also define the query-dependent table representation
\[
\mathbf{r}_t = \mathbf{v}^{*}_t,
\qquad
v_t^* =
\arg\max_{v \in V(t)}
\langle \mathbf{q}, \mathbf{v} \rangle ,
\label{eq:table-representation}
\]
where $\mathbf{v}^{*}_t$ is the embedding of the best-matching node $v_t^*$ in table $t$.
The top-scoring tables are selected as anchors for graph traversal.

\subsubsection{Learned Graph Walk}
\label{sec:walk}

From each anchor table, \joingr performs a bounded breadth-first walk of depth $H$
over join edges. The walk maintains a non-negative deposit at each frontier
table, representing the amount of relevance that has reached that table. At
each hop, the model distributes this deposit across the outgoing join edges
according to query-conditioned transition scores.

Let $f$ denote the frontier table at hop $h$, with incoming deposit $z_f$.
For each outgoing join edge $e=(f,t')$, where $t'$ is the neighboring table, we
form the feature vector
\[
  \mathbf{x}_e
  =
  [\,\mathbf{q}\,;\,\mathbf{e}\,;\,\mathbf{r}_{f}\,;\,\mathbf{r}_{t'}\,;\,z_f\,;\,h\,],
  \label{eq:feat}
\]
where $\mathbf{q}$ is the query embedding, $\mathbf{e}$ is the edge embedding, $\mathbf{r}_{f}$ and $\mathbf{r}_{t'}$ are the query-dependent table
representations of the frontier and neighbor tables, respectively, $z_f$ is the
incoming deposit, and $h$ is the hop index.

\begin{figure}[t]
\centering
\resizebox{\columnwidth}{!}{%
\begin{tikzpicture}[
  node/.style={
    circle,
    draw,
    thick,
    minimum size=1.5cm,
    align=center,
    font=\small
  },
  frontier/.style={
    circle,
    draw,
    thick,
    minimum size=1.5cm,
    align=center,
    font=\large,
    fill=orange!18
  },
  nbr/.style={
    circle,
    draw,
    thick,
    minimum size=1.5cm,
    align=center,
    font=\large,
    fill=blue!10
  },
  edge/.style={->, >=stealth},
  lab/.style={font=\large, fill=white, inner sep=1pt}
]

\node[frontier] (f) at (1,0) {$f$};

\node[nbr] (t1) at (3.0,1.55) {$t_1$};
\node[nbr] (t2) at (5.0,0) {$t_2$};
\node[nbr] (t3) at (3.0,-1.55) {$t_3$};

\node (r1) at (6.0, 2.2) {};
\node (r3) at (6.2, -1.2) {};

\draw[edge, purple, line width=0.5pt, bend left=12]
  (f) to (t1);

\draw[edge, purple, line width=2.0pt]
  (f) to node[above, lab] {$e:\ u_{e}\!\to\!\pi_{e}$} (t2);

\draw[edge, purple, line width=1.2pt, bend right=12]
  (f) to (t3);

\draw[edge, purple, dashed, line width=1.2pt, bend left=12] (t2) to (r1);
\draw[edge, purple, dashed, line width=.7pt, bend right=12] (t2) to (r3);

\draw[->, dashed, purple, line width=1.2pt]
  (-0.5,1.0) to[bend right=20] node[above, lab] {$z_f$} (f.west);

\node[font=\large, align=center, purple] at (3.0,-0.3)
{$\Delta_{e}=z_f\alpha_h\pi_{e}$};

\end{tikzpicture}%
}
\caption{Expansion in Join Graph. From frontier table $f$, \joingr scores outgoing join edges to neighbor $t_1$, $t_2$ and $t_3$, locally normalizes the scores, and propagates deposit.}
\label{fig:learned_walk}
\end{figure}
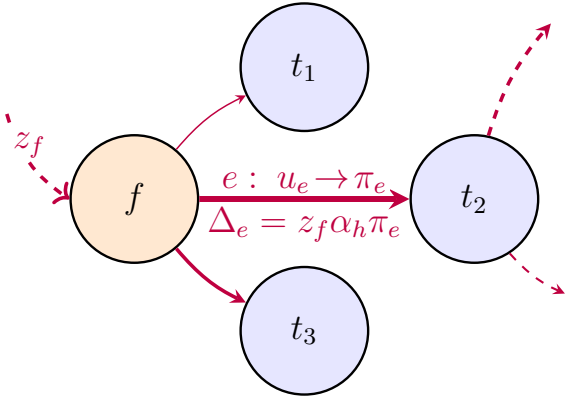

A query-conditioned scorer $s_\theta$ assigns a relevance score
to each outgoing edge:
\[
  u_e = s_\theta(\mathbf{x}_e) \in \mathbb{R}.
  \label{eq:scorer}
\]
For each frontier table $f$, the raw scores of its outgoing join edges $\mathcal{N}_{\mathrm{join}}(f)$ are locally normalized with a softmax:
\[
  \pi_e
  =
  \frac{\exp(u_e)}
  {\sum_{e' \in \mathcal{N}_{\mathrm{join}}(f)} \exp(u_{e'})}.
  \label{eq:transition}
\]
The deposit propagated onto edge $e$ is
\[
  \Delta_e = z_f \cdot \alpha_h \cdot \pi_e,
  \label{eq:deposit}
\]
where $\alpha_h=\sigma(\tilde{\alpha}_h)$ is a learned hop-specific decay coefficient. Figure~\ref{fig:learned_walk} summarizes this local expansion step.

Edge deposits accumulate across all anchors and hops. If the neighboring table $t'$ has not already been expanded in the current anchor walk, $\Delta_e$
is added to its deposit for the next hop. We maintain a visited set for each anchor walk to prevent cycles.


\subsubsection{Aggregation}
\label{sec:aggregation}

The graph walk produces edge-level evidence, but the retrieval output is a
ranked list of tables. We therefore project accumulated edge deposits back onto
their endpoint tables. Let $\Delta_e(q)$ denote the total deposit accumulated
on edge $e$ across all anchors and hops. The final table score is
\[
  \hat{s}(t \mid q)
  =
  \sum_{e:\, t \in \mathrm{endpoints}(e)}
  \Delta_e(q).
  \label{eq:aggregation}
\]
Tables are ranked in descending order of $\hat{s}(t \mid q)$.



\subsubsection{Training}
\label{sec:training}

We train the edge scorer using supervised table sets. Let
$P=\mathcal{T}^g(q)$ be the gold table set and
$N=\mathcal{T}(D)\setminus P$ the negatives.
The \emph{table-level} loss is
\[
  \mathcal{L}_{\text{tab}} \;=\;
  \frac{1}{|P||N|}
  \sum_{\substack{p\in P\\n\in N}}
  \max\!\big(0,\,m - \hat{s}(p) + \hat{s}(n)\big),
  \label{eq:loss}
\]
with margin $m$.

When gold join keys $\mathcal{K}^g(q)$ are available, we also consider an edge-level variant that applies the same pairwise margin objective to deposits on gold and non-gold join edges. The main experiments use the table-level loss, since gold table sets are available in standard text-to-SQL benchmarks.



\section{Experimental Setup}
\label{sec:setup}

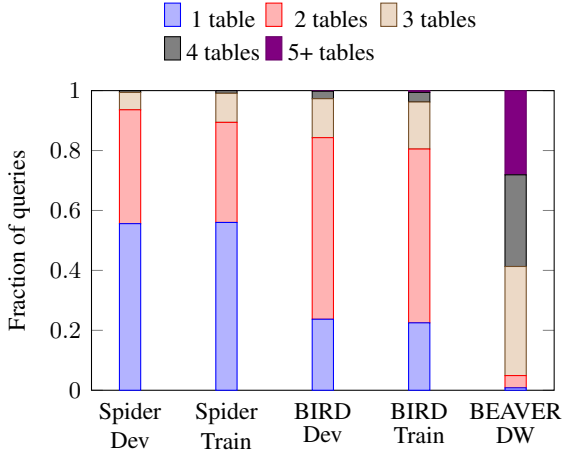
\begin{figure}[ht]
\centering
\begin{tikzpicture}
\begin{axis}[
    ybar stacked,
    width=\columnwidth,
    height=0.72\columnwidth,
    bar width=8pt,
    ylabel={Fraction of queries},
    symbolic x coords={Spider-D,Spider-T,BIRD-D,BIRD-T,BEAVER},
    xtick=data,
    xticklabels={
        {\shortstack{Spider\\Dev}},
        {\shortstack{Spider\\Train}},
        {\shortstack{BIRD\\Dev}},
        {\shortstack{BIRD\\Train}},
        {\shortstack{BEAVER\\DW}}
    },
    x tick label style={rotate=0, align=center, yshift=-1pt},
    ymin=0,
    ymax=1,
    legend style={
        at={(0.5,1.05)},
        anchor=south,
        legend columns=3,
        font=\small,
        draw=none
    },
    tick label style={font=\small},
    label style={font=\small},
]
\addplot coordinates {(Spider-D,0.556) (Spider-T,0.560) (BIRD-D,0.237) (BIRD-T,0.225) (BEAVER,0.008)};
\addplot coordinates {(Spider-D,0.380) (Spider-T,0.334) (BIRD-D,0.606) (BIRD-T,0.580) (BEAVER,0.041)};
\addplot coordinates {(Spider-D,0.058) (Spider-T,0.097) (BIRD-D,0.130) (BIRD-T,0.157) (BEAVER,0.364)};
\addplot coordinates {(Spider-D,0.006) (Spider-T,0.009) (BIRD-D,0.025) (BIRD-T,0.032) (BEAVER,0.306)};
\addplot coordinates {(Spider-D,0.000) (Spider-T,0.001) (BIRD-D,0.002) (BIRD-T,0.006) (BEAVER,0.281)};
\legend{1 table,2 tables,3 tables,4 tables,5+ tables}
\end{axis}
\end{tikzpicture}
\caption{Distribution of queries by number of required gold tables.}
\label{fig:gold-table-dist}
\end{figure}

\paragraph{Datasets.}
We evaluate \joingr on three text-to-SQL benchmarks that offers varying degrees of table-retrieval complexity: Spider \cite{yu2018spider}, BIRD \cite{li2023bird} and BEAVER \cite{stonebraker2024beaver}. Table~\ref{tab:dataset-graph-stats} summarizes dataset and join-graph statistics for each split. Figure~\ref{fig:gold-table-dist} shows the distribution of the number of gold tables required per query.

\begin{table}[ht]
\centering
\small
\setlength{\tabcolsep}{2.5pt}
\begin{tabular}{llrrrrrr}
\toprule
Dataset & Split &\#DB & \#Tab. & \#Q & \#N & \#Intra & \#Join \\
\midrule
Spider & Dev   & 20  & 81  & 1034 & 441   & 492   & 63  \\
Spider & Train & 166 & 876 & 7000 & 4503 & 4704 & 793 \\
BIRD   & Dev   & 11  & 75  & 1534 & 798   & 3345 & 102 \\
BIRD   & Train & 69  & 522 & 9428 & 3539 & 4559 & 418 \\
BEAVER & DW    & 1   & 97  & 121   & 1530 & 3536 & 197 \\
\bottomrule
\end{tabular}
\caption{Dataset and join-graph statistics. \#N denotes column nodes; intra edges connect columns within a table, and join edges connect foreign-key-related columns across tables.}
\label{tab:dataset-graph-stats}
\end{table}

\paragraph{Encoders and baselines.}
We use two frozen sentence embedders \cite{reimers2019sbert}, all-MiniLM-L6-v2 \cite{wang2020minilm} and BGE-large-en-v1.5 \cite{xiao2024cpack} to embed queries, nodes, and edges. Our first baseline is the corresponding \emph{Vanilla Retriever}, which ranks each table by the maximum similarity between the query and any column in the table. This baseline isolates semantic retrieval without graph expansion.

We compare against recent multi-table retrieval baselines. REaR \cite{agarwal2025rear} is a retrieve-and-rerank pipeline that refines initially retrieved candidates using relation-aware signals. JAR \cite{chen-etal-2024-table} is a join-aware re-ranking method that combines query-table relevance with table-table joinability and solves table selection as a mixed-integer optimization problem. IterativeJAR \cite{boutaleb2025exploring} replaces JAR's global optimization step with a greedy iterative search procedure for better scalability. We report the available DTR- and Contriever-based variants of these methods where applicable.

Finally, we include a heuristic \joingr variant that uses the same query-conditioned anchor--walk--aggregate pipeline but replaces the learned edge scorer with fixed propagation weights. Comparing dense retrieval, heuristic \joingr, and trained \joingr isolates the contributions of semantic matching, graph expansion with fixed weights, and learned edge weighting.


\paragraph{Scorer implementation and training.}
The \joingr scorer is a one-hidden-layer MLP with LayerNorm, ReLU, and dropout, mapping the edge feature vector described in Section~\ref{sec:walk} to an unnormalized edge score. We use hidden dimension 64, dropout 0.3, Adam with learning rate $10^{-3}$ and weight decay $10^{-4}$, gradient clipping at 1.0, and ReduceLROnPlateau scheduling. We set maximum walk depth $H=5$. The sentence encoder and all graph embeddings are frozen; only the edge scorer and hop-specific decay parameters are trained.

Spider and BIRD use the standard train/dev split. Because BEAVER (DW) has only 121 queries, we report 5-fold cross-validation. Cross-domain experiments train on one benchmark and evaluate on another without target-domain supervision.

\paragraph{Metrics.}
We report precision (P@K), recall (R@K), and perfect recall
(PR@K) at $K\in\{5,15\}$.
PR@K is the strictest metric: it equals $1$ iff every gold table is present within the first $K$ elements, and is the metric most directly tied to downstream SQL accuracy.


\section{Results}
\label{sec:results}
\subsection{Results on BEAVER (DW)}
\label{sec:beaver-results}
Table~\ref{tab:beaver_results} reports results on BEAVER (DW), the most structurally challenging benchmark in our evaluation. Across both encoders, \joingr{} achieves the strongest overall retrieval performance. With BGE-large-en-v1.5, \joingr{} improves R@15 from 0.697 for the vanilla retriever to 0.875, and PR@15 from 0.339 to 0.645. With MiniLM, the gains are similarly large, improving R@15 from 0.667 to 0.873 and PR@15 from 0.273 to 0.653.

The gains are especially visible with the lightweight MiniLM encoder, where semantic retrieval is weaker. For example, \joingr{} improves R@5 from 0.385 to 0.528 and PR@5 from 0.066 to 0.182 over the vanilla retriever. This suggests that structural traversal can compensate when embedding quality alone is insufficient, reducing the dependence on stronger semantic encoders.

Because BEAVER (DW) contains only 121 annotated queries, we report the in-domain trained model using 5-fold cross-validation. This limited evaluation size leads to noticeable fold-to-fold variance, so we interpret the in-domain numbers as evidence of the method's potential under supervision. The cross-domain setting provides an important complementary evaluation. A scorer trained on BIRD and evaluated on BEAVER (DW) still achieves R@15 of 0.806 with BGE and 0.740 with MiniLM, matching other retrieval and re-ranking baselines. This suggests that the scorer learns a reusable scoring mechanism over join graphs.

The comparison between the vanilla retriever, heuristic \joingr{}, and trained \joingr{} further isolates the contributions of semantic similarity, graph expansion, and learned edge scoring. The heuristic variant improves R@15 over the vanilla retriever for both encoders, indicating that join-graph expansion alone recovers additional relevant tables. However, trained \joingr{} substantially improves perfect recall over the heuristic variant, increasing PR@15 from 0.405 to 0.645 with BGE and from 0.430 to 0.653 with MiniLM. This indicates that learning how to distribute deposit is important for recovering complete multi-hop table sets.

\begin{table}[ht]
\centering
\scriptsize
\setlength{\tabcolsep}{5pt}
\begin{tabular}{l|cc|cc|cc}
\toprule
\multirow{2}{*}{Method} &
\multicolumn{2}{c|}{P} &
\multicolumn{2}{c|}{R} &
\multicolumn{2}{c}{PR} \\
& @5 & @15 & @5 & @15 & @5 & @15 \\
\midrule

\multicolumn{7}{c}{\textbf{BGE-large-en-v1.5}} \\
\midrule

Vanila Retriever
& 0.352 & 0.175
& 0.476 & 0.697
& 0.124 & 0.339 \\

REaR
& 0.347 & 0.179
& 0.474 & 0.684
& 0.174 & 0.364 \\

\joingr{}$_{\text{H}}$
& 0.319 & 0.195
& 0.423 & 0.759
& 0.058 & 0.405 \\

\textbf{\joingr{}}
& \textbf{0.423} & \textbf{0.226}
& \textbf{0.553} & \textbf{0.875}
& 0.157 & \textbf{0.645} \\

\joingr{}{$_{\text{X}\rightarrow \text{Y}}$}
& 0.384 & 0.206
& 0.514 & 0.806
& 0.149 & 0.529 \\

\midrule
\multicolumn{7}{c}{\textbf{all-MiniLM-L6-v2}} \\
\midrule

Vanilla Retriever
& 0.279 & 0.169
& 0.385 & 0.667
& 0.066 & 0.273 \\

REaR
& 0.247 & 0.134
& 0.340 & 0.445
& 0.116 & 0.215 \\

\joingr{}$_{\text{H}}$ 
& 0.311 & 0.200
& 0.414 & 0.779
& 0.074 & 0.430 \\

\textbf{\joingr{}}
& \textbf{0.395} & \textbf{0.226}
& \textbf{0.528} & \textbf{0.873}
& \textbf{0.182} & \textbf{0.653} \\

\joingr{}{$_{\text{X}\rightarrow \text{Y}}$}
& 0.314 & 0.187
& 0.429 & 0.740
& 0.107 & 0.397 \\

\midrule
\multicolumn{7}{c}{\textbf{Additional Baselines}} \\
\midrule

DTR
& 0.323 & 0.176
& 0.431 & 0.693
& 0.150 & 0.375 \\

Contriever
& 0.353 & 0.176
& 0.468 & 0.695
& 0.142 & 0.392 \\

JAR$_{\text{F (DTR)}}$ 
& 0.113 & --
& 0.155 & --
& 0.067 & -- \\

JAR$_{\text{F (Contriever)}}$ 
& 0.303 & --
& 0.392 & --
& 0.125 & -- \\

JAR$_{\text{Iterative}}$
& 0.379 & 0.180
& 0.515 & 0.719
& 0.198 & 0.397 \\

\bottomrule
\end{tabular}

\caption{
Table retrieval results on \textbf{BEAVER (DW)}. \joingr{}$_{\text{H}}$ is the heuristic version of our algorithm. For \joingr{}{$_{\text{X}\rightarrow \text{Y}}$} we utilized the scorer trained on BIRD dataset and evaluated on BEAVER.
}
\label{tab:beaver_results}
\end{table}

\subsection{Results on BIRD}
\label{sec:bird-results}
Table~\ref{tab:bird_results} reports results on BIRD. In comparison with BEAVER, BIRD is a high-recall setting for dense retrieval: 84\% queries require only one or two tables. As a result, the vanilla retriever is already strong, reaching R@15 of 0.983 and PR@15 of 0.964 with BGE.

In this setting, \joingr{} remains competitive. With BGE, REaR achieves the strongest R@15 and PR@15, while \joingr{} obtains the best P@5 and PR@5. The heuristic variant underperforms, suggesting that fixed-weight graph expansion can over-explore when semantic retrieval already identifies most relevant tables.

The benefit of learned traversal is clearer with the lightweight MiniLM encoder. Compared with the MiniLM vanilla retriever, \joingr{} improves R@5 from 0.886 to 0.909 and PR@15 from 0.936 to 0.951. This mirrors the trend observed on BEAVER: when the semantic retriever is weaker, query-conditioned join traversal provides a useful corrective signal.

Alignment-based re-rankers are also strong on BIRD: REaR performs query-aware cross-encoder reranking in its refine stage, while JAR and IterativeJAR incorporate fine-grained query--schema alignment and coverage-based selection. These signals are well suited to BIRD, where many required tables are directly reflected in the utterance.


\begin{table}[!ht]
\centering
\scriptsize
\setlength{\tabcolsep}{5pt}
\begin{tabular}{l|cc|cc|cc}
\toprule
\multirow{2}{*}{Method} &
\multicolumn{2}{c|}{P} &
\multicolumn{2}{c|}{R} &
\multicolumn{2}{c}{PR} \\
& @5 & @15 & @5 & @15 & @5 & @15 \\
\midrule

\multicolumn{7}{c}{\textbf{BGE-large-en-v1.5}} \\
\midrule

Vanilla Retriever
& 0.349 & 0.127
& 0.918 & 0.983
& 0.830 & 0.964 \\

REaR
& 0.359 & \textbf{0.198}
& \textbf{0.937} & \textbf{0.985}
& 0.871 & \textbf{0.970} \\

\joingr{}$_{\text{H}}$
& 0.300 & 0.123
& 0.789 & 0.951
& 0.650 & 0.921 \\

\textbf{\joingr{}}
& \textbf{0.360} & 0.188
& 0.934 & 0.973
& \textbf{0.877} & 0.961 \\

\joingr{}$_{\text{X}\rightarrow \text{Y}}$
& 0.346 & 0.126
& 0.901 & 0.973
& 0.818 & 0.962 \\

\midrule
\multicolumn{7}{c}{\textbf{all-MiniLM-L6-v2}} \\
\midrule

Vanilla Retriever
& 0.335 & 0.125
& 0.886 & \textbf{0.970}
& 0.775 & 0.936 \\

REaR
& \textbf{0.354} & \textbf{0.189}
& \textbf{0.923} & 0.969
& \textbf{0.846} & 0.942 \\

\joingr{}$_{\text{H}}$
& 0.325 & 0.122
& 0.858 & 0.945
& 0.762 & 0.917 \\

\textbf{\joingr{}}
& 0.347 & 0.125
& 0.909 & 0.966
& 0.839 & \textbf{0.951} \\

\joingr{}$_{\text{X}\rightarrow \text{Y}}$
& 0.338 & 0.125
& 0.883 & 0.966
& 0.787 & \textbf{0.951} \\

\midrule
\multicolumn{7}{c}{\textbf{Additional Baselines}} \\
\midrule

DTR
& 0.376 & 0.144
& 0.851 & 0.970
& 0.701 & 0.934 \\

Contriever
& 0.366 & 0.145
& 0.825 & 0.974
& 0.654 & 0.949 \\

JAR$_{\text{F (DTR)}}$
& 0.402 & --
& 0.897 & --
& -- & -- \\

JAR$_{\text{F (Contriever)}}$
& 0.403 & --
& 0.898 & --
& -- & -- \\

JAR$_{\text{Iterative}}$
& 0.355 & 0.127
& 0.930 & 0.980
& 0.864 & 0.961 \\

\bottomrule
\end{tabular}

\caption{
Table retrieval results on \textbf{BIRD}. \joingr{}$_{\text{H}}$ is the heuristic version of our algorithm. For \joingr{}{$_{\text{X}\rightarrow \text{Y}}$} we utilized the scorer trained on BEAVER dataset and evaluated on BIRD.
}
\label{tab:bird_results}
\end{table}

\subsection{Results on SPIDER}
\label{sec:spider-results}
Spider is the least structurally demanding benchmark in our evaluation: 94\% of queries require only one or two tables, and the schemas are well structured. For this dataset, retrieval is largely solved by direct query-schema similarity. Consequently, \joingr{} remains close to the corresponding vanilla retriever. With MiniLM, \joingr{} achieves R@15 of 0.748 compared with 0.752 for the vanilla retriever, indicating that query-conditioned traversal does not disrupt simple retrieval. Existing alignment-based baselines such as DTR, Contriever, JAR, and IterativeJAR perform strongly on Spider, reflecting that many required tables are directly recoverable from the query utterance. Full Spider results are reported in Appendix~\ref{sec:appendix-spider}




\subsection{Cross-Domain Transfer}
\label{sec:cross}
\label{sec:transfer}

The cross-domain rows, \joingr{}$_{\text{X}\rightarrow \text{Y}}$, in Tables~\ref{tab:beaver_results}, \ref{tab:bird_results}, and \ref{tab:spider_results} evaluate transfer without target-domain supervision: the scorer is trained on dataset X and evaluated on dataset Y. This is crucial when annotated examples are scarce, as a scorer from another domain can serve as an initial tool until local supervision is collected.
\newline\indent Section~\ref{sec:beaver-results} shows BEAVER (DW) results as strong evidence: despite limited supervision, the BIRD-trained scorer is competitive with in-domain \joingr{} and other retrieval baselines, indicating that the scorer learns reusable behavior rather than fitting only the source benchmark.
\newline\indent Transfer is less critical on BIRD and Spider, where vanilla retrieval is strong. However, BEAVER-trained scorers remain close to in-domain \joingr{} on these datasets. Overall, cross-domain results support using \joingr{} as a reusable scoring mechanism that can be deployed before domain-specific supervision and then fine-tuned when local data justifies a domain-specific scorer.
\section{Analysis}
\label{sec:analysis}
\subsection{Precision and Recall by Query Complexity}
\begin{figure}[!htbp]
\centering

\hspace*{-0.05\columnwidth}
\begin{tikzpicture}
\begin{axis}[
    ybar,
    width=1.08\columnwidth,
    height=0.62\columnwidth,
    bar width=3.2pt,
    ymin=0,
    ymax=1.25,
    ylabel={},
    xmin=0.2,
    xmax=14.8,
    xtick={1,2,3,4,5,6,9,10,11,12,13,14},
    xticklabels={},
    tick style={draw=none},
    x tick label style={font=\scriptsize},
    yticklabels={},
    ytick style={draw=none},
    tick label style={font=\scriptsize},
    label style={font=\scriptsize},
    nodes near coords,
    nodes near coords style={
        font=\scriptsize,
        rotate=90,
        anchor=west
    },
    every node near coord/.append style={
        /pgf/number format/fixed,
        /pgf/number format/precision=2,
    },
]

\addplot coordinates {
    (1,0.347) (2,0.190) (3,0.371) (4,0.486) (5,0.532) (6,0.533)
    (9,0.125) (10,0.065) (11,0.130) (12,0.188) (13,0.256) (14,0.244)
};

\addplot coordinates {
    (1,0.909) (2,0.950) (3,0.927) (4,0.810) (5,0.665) (6,0.500)
    (9,0.966) (10,0.967) (11,0.974) (12,0.940) (13,0.961) (14,0.689)
};

\draw[dashed, gray] (axis cs:7.5,0) -- (axis cs:7.5,1.1);

\node[font=\small] at (rel axis cs:0.5,.93) {(a)};
\end{axis}
\end{tikzpicture}

\vspace{-0.15em}

\hspace*{-0.05\columnwidth}
\begin{tikzpicture}
\begin{axis}[
    ybar,
    width=1.08\columnwidth,
    height=0.62\columnwidth,
    bar width=3.2pt,
    ymin=0,
    ymax=1.15,
    ylabel={},
    xmin=0.2,
    xmax=14.8,
    xtick={1,2,3,4,5,6,9,10,11,12,13,14},
    xticklabels={ALL,1,2,3,4,5+,ALL,1,2,3,4,5+},
    tick style={draw=none},
    x tick label style={font=\scriptsize},
    yticklabels={},
    ytick style={draw=none},
    extra x ticks={3.5,11.5},
    extra x tick labels={@5,@15},
    extra x tick style={
        tick style={draw=none},
        grid=none,
        xticklabel style={font=\scriptsize, yshift=-14pt}
    },
    tick label style={font=\scriptsize},
    label style={font=\scriptsize},
    nodes near coords,
    nodes near coords style={
        font=\scriptsize,
        rotate=90,
        anchor=west
    },
    every node near coord/.append style={
        /pgf/number format/fixed,
        /pgf/number format/precision=2,
    },
]

\addplot coordinates {
    (1,0.314) (2,0.200) (3,0.280) (4,0.314) (5,0.297) (6,0.341)
    (9,0.187) (10,0.067) (11,0.120) (12,0.171) (13,0.177) (14,0.231)
};

\addplot coordinates {
    (1,0.429) (2,1.000) (3,0.700) (4,0.523) (5,0.372) (6,0.313)
    (9,0.740) (10,1.000) (11,0.900) (12,0.856) (13,0.662) (14,0.643)
};

\draw[dashed, gray] (axis cs:7.5,0) -- (axis cs:7.5,1.0);

\node[font=\small] at (rel axis cs:0.5,.93) {(b)};
\end{axis}
\end{tikzpicture}

\vspace{0.2em}

\begin{tikzpicture}
\begin{axis}[
    hide axis,
    xmin=0, xmax=1,
    ymin=0, ymax=1,
    legend columns=2,
    legend style={
        draw=none,
        font=\scriptsize,
        column sep=1em
    },
    legend image code/.code={
        \draw[#1, fill] (0cm,-0.05cm) rectangle (0.16cm,0.11cm);
    }
]

\addlegendimage{blue, fill=blue}
\addlegendentry{Precision}

\addlegendimage{red, fill=red}
\addlegendentry{Recall}

\end{axis}
\end{tikzpicture}

\caption{Precision and Recall @5, 15 by number of gold tables: (a) BIRD, (b) BEAVER (DW).}
\label{fig:complexity_pr}
\end{figure}
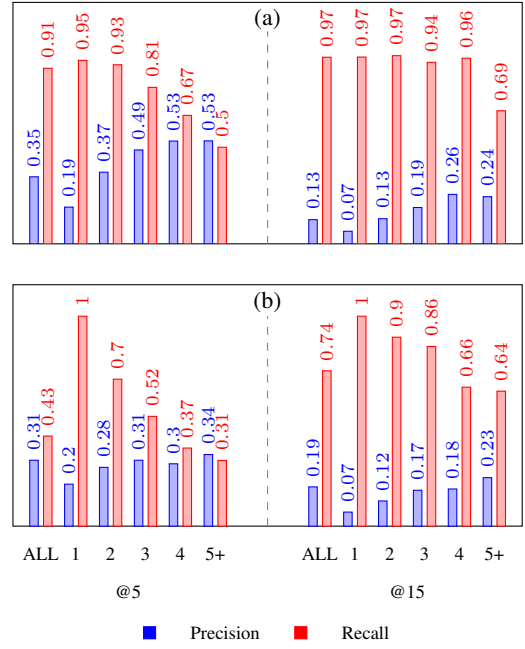

We analyze precision and recall by query complexity, using the number of required gold tables as a proxy for the join structure needed. Queries are bucketed into \textsc{All}, 1, 2, 3, 4, and 5+ required tables, reporting P/R at @5 and @15 for MiniLM-\joingr{}. We focus on the BIRD-trained scorer in two settings: BIRD (Dev), with many queries answerable from one or two tables, and BEAVER (DW), where queries often need multi-hop relational context. The BIRD scorer on BEAVER avoids evaluating queries possibly seen during BEAVER cross-validation.
\newline\indent Figure~\ref{fig:complexity_pr} shows traversal preserves performance on simpler BIRD queries: for one- and two-table buckets, recall remains high, indicating the learned walk doesn't disrupt cases where semantic retrieval identifies needed tables. BEAVER (DW) buckets show why traversal is necessary for harder queries. As required table count grows, recall remains higher, showing the walk recovers relevant tables even when retrieval becomes harder. The drop for 4- and 5+-table BEAVER queries reflects the difficulty of following longer join paths without target-domain supervision.

\subsection{Semantic Anchors vs. Graph Expansion}
We isolate the value of structured traversal by varying the number of anchor tables on BEAVER (DW). All methods use MiniLM and are evaluated at $K=10$. We use R@10 because BEAVER (DW) queries can require up to seven tables. 

The Vanilla Retriever returns tables directly from the semantic ranking. In contrast, \joingr{} starts from the same anchors but expands through the join graph, allowing it to retrieve additional tables.

\begin{figure}[ht]
\centering
\begin{tikzpicture}
\begin{axis}[
    width=\columnwidth,
    height=0.62\columnwidth,
    xlabel={Number of anchors},
    ylabel={R@10},
    xmin=1,
    xmax=7,
    ymin=0,
    ymax=0.70,
    xtick={1,2,3,4,5,6,7},
    tick label style={font=\scriptsize},
    label style={font=\scriptsize},
    legend style={
        at={(0.5,1.05)},
        anchor=south,
        legend columns=3,
        font=\scriptsize,
        draw=none
    },
    mark size=1.8pt,
]
\addplot+[mark=*] coordinates {
    (1,0.123) (2,0.205) (3,0.265) (4,0.314) (5,0.363) (6,0.415) (7,0.449)
};
\addplot+[mark=square*] coordinates {
    (1,0.493) (2,0.584) (3,0.610) (4,0.630) (5,0.604) (6,0.611) (7,0.609)
};
\addplot+[mark=triangle*] coordinates {
    (1,0.496) (2,0.577) (3,0.610) (4,0.621) (5,0.621) (6,0.639) (7,0.634)
};
\legend{Vanilla, Heuristic, Trained}
\end{axis}
\end{tikzpicture}
\caption{R@10 on BEAVER (DW) for different anchor budgets.}
\label{fig:semantic-structured}
\end{figure}
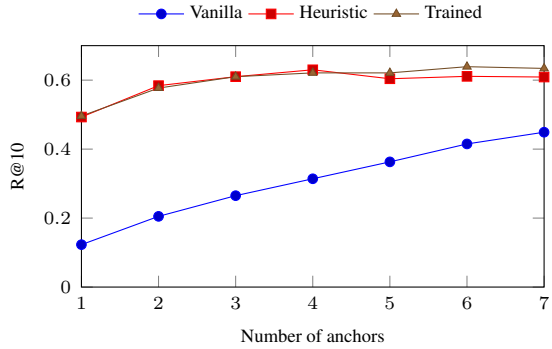

Figure~\ref{fig:semantic-structured} compares the Vanilla Retriever, heuristic \joingr{}, and trained \joingr{} as the number of anchors increases from 1 to 7. The gap between Vanilla and the graph-based variants is largest when the anchor budget is small. This indicates that many relevant BEAVER (DW) tables are not among the top semantic matches to the query, but can be recovered once retrieval expands from a related anchor through schema structure.

As the number of anchors grows, Vanilla recall improves because more semantically similar tables are included directly, but it remains below the graph-based methods. The heuristic variant captures much of the gain from adding structure, while the trained scorer provides the best overall recall, especially at larger anchor budgets. This supports the central design of \joingr{}: the join graph supplies the expansion space, and learned edge weighting improves how relevance is distributed through that space.

\subsection{Failure Cases}
We inspect the BEAVER queries on which Trained-\joingr misses at least one
gold table at $K{=}15$.
The dominant failure mode is the long tail of ``narrative'' joins:
questions like \textit{``Which department occupies the most square
footage?''} require traversing four FK chains, and the third hop's softmax
frequently splits mass across two plausible neighbours.
A second failure mode is anchor mis-selection: when the question vocabulary
overlaps several tables (e.g., \texttt{building}, \texttt{address},
\texttt{room}), the top-$5$ anchors fail to include the eventual main table.
Both failure modes are mitigated through additional anchors or hops, at the cost of
latency.




\section{Related Work}
\label{sec:related}

\paragraph{Table retrieval and schema linking for text-to-SQL.}
Text-to-SQL systems identify relevant schema elements before generation, either within the parser or as an explicit upstream step~\cite{wang2020ratsql,guo2019irnet,pourreza2023din}. Parser- and prompting-based systems assume the database schema for a question is provided, then perform schema linking. Recent LLM-based pipelines make this step crucial as irrelevant schema elements can distract generation or cause spurious joins~\cite{maamari2024the}. While dense retrievers and prompt-based linkers rely on query to schema similarity, \joingr{} adds join-graph traversal, considering structurally related tables during retrieval.

\paragraph{Table representations and join-aware retrieval.}
Table encoders~\cite{herzig2021dtr, deng2022turl}, and dense retrievers~\cite{izacard2022contriever}, learn semantic representations of tables, columns, or their serialized forms. Join-aware methods~\cite{chen-etal-2024-table, boutaleb2025exploring, agarwal2025rear}  add structural signals during retrieval or re-ranking. \joingr{} differs by using the database join graph itself as the retrieval space.

\paragraph{Graph-based retrieval.}
Graph retrieval has been widely used in multi-hop QA and document retrieval~\cite{zhang-etal-2022-subgraph, gutierrez2024hipporag, gutierrez2025from}, where systems select anchors and expand through edges to recover evidence spread across multiple units. \joingr{} adapts this idea to relational schemas, where columns and tables form a natural hierarchy and foreign-key edges define possible retrieval paths. 

\paragraph{Schema graphs for tabular reasoning.}
Graph-based methods also appear for multi-table reasoning. \cite{wang-etal-2025-plugging} uses schema graphs for end-to-end QA/SQL generation, while \cite{safdarian-etal-2026-schemagraphsql} performs training-free schema linking by predicting tables and returning shortest paths in a table-level graph. \joingr{} focuses on upstream table retrieval, learning a traversal scorer over a column-level join graph, allowing retrieval through query-supported join paths.

Concurrent anonymous work, CORE-T \cite{annonymous_coret} tackles open-book multi-table retrieval for text-to-SQL on pooled heterogeneous tables without database IDs or gold foreign keys. It integrates offline LLM-generated metadata, a compatibility cache, and online dense retrieval with LLM selection to produce compact, join-coherent table sets. Whereas, \joingr{} uses a learned scorer for join-graph traversal.

\section{Conclusion}
\label{sec:conclusion}
We present \joingr{}, a table retriever using the database join graph as a search space. Retrieval starts from semantic anchors, follows join edges with a query-conditioned scorer, and aggregates evidence into table scores. This design separates semantic matching from structural expansion, allowing use of both direct question matches and schema connections. The main takeaway is that join structure guides retrieval, not just SQL generation. In settings where independent table similarity suffices, \joingr{} competes with standard retrievers. When tables connect through relational context, learned traversal aids recovery. \joingr{} exemplifies the principle: \emph{when information connects by structure, retrieval should use it}.
\section{Limitations}
\label{sec:limitations}

\joingr depends on a join graph being constructable from the schema.
For databases that lack declared foreign-key constraints, foreign keys must
be inferred, typically via heuristic name-matching or an LLM pass, and the
noise of that inference propagates into our edges.
Second, our experiments use frozen encoders; we have not measured the effect of fine-tuning the encoder jointly with the scorer, and we expect
that to be a useful but tangential extension.
Third, the LLM-generated descriptors used in node and edge text are a one-time cost that scales linearly in the number of columns plus edges; on
a $10{,}000$-column schema this is tractable but not free. Finally, our evaluation is restricted to English-language questions and to three benchmarks; multilingual and lower-resource settings remain unexplored.

\section{Ethical Considerations}
\label{sec:ethics}

Table Retrieval is an enabling step for natural-language access to enterprise
databases, with corresponding privacy implications: incorrect retrieval can
disclose tables the user is not authorized to see, while correct retrieval
enables them to query data that may be sensitive.
Our method is no more or less prone to either failure than the baselines it
compares against; we recommend that any deployment compose \joingr with a
downstream authorization layer that is independent of the retrieval policy.
We use LLMs only to generate schema descriptions, on data that the schema
owner controls; no personal data leaves the database. AI writing assistants were used to polish the manuscript; all scientific claims, experimental design, and analysis were conducted and verified by the authors.

\bibliography{custom}
\clearpage
\appendix

\section{Algorithms}
\label{sec:algorithms}
Algorithm~\ref{alg:graph_construction} constructs a navigation graph over all databases in the dataset. Algorithm~\ref{alg:search} performs query-conditioned graph traversal using offline semantic indexes and a learned edge scorer. Starting from retrieved anchor tables, the algorithm propagates relevance scores through join paths and ranks tables based on accumulated traversal deposits.
\begin{algorithm}[!ht]
\caption{\small Abstract Join Graph Construction}
\label{alg:graph_construction}
\scriptsize
\begin{algorithmic}[1]

\Require Tables $\mathcal{T} = \{T_1, T_2, \dots, T_m\}$

\ForAll{table $T \in \mathcal{T}$}
    \State $C(T)$: set of columns in $T$
    \State $K(T) \subseteq C(T)$: set of key columns
\EndFor

\Require $\mathcal{J}$: set of join relationships between key columns
\Ensure Graph $G=(V,E)$

\State $V \gets \emptyset,\; E \gets \emptyset$

\Statex
\Comment{Add column nodes}
\ForAll{table $T \in \mathcal{T}$}
    \ForAll{column $c \in C(T)$}
        \State Add node $v(c)$ to $V$
        \State Mark whether $c \in K(T)$
    \EndFor
\EndFor

\Statex
\Comment{Add intra-table navigation edges}
\ForAll{table $T \in \mathcal{T}$}

    \State $N(T) \gets C(T) \setminus K(T)$

    \Comment{Connect non-key columns to key columns}
    \ForAll{$n \in N(T)$}
        \ForAll{$k \in K(T)$}
            \State Add edge $(v(n), v(k))$ to $E$ with type \texttt{intra}
        \EndFor
    \EndFor

    \Comment{Connect key columns pairwise}
    \ForAll{unordered pair $(k_i, k_j) \in K(T)$}
        \State Add edge $(v(k_i), v(k_j))$ to $E$ with type \texttt{intra}
    \EndFor

\EndFor

\Statex
\Comment{Add inter-table join edges}
\ForAll{join relationship between $k_a \in K(T_a)$ and $k_b \in K(T_b)$}
    \State Add edge $(v(k_a), v(k_b))$ to $E$ with type \texttt{join}
\EndFor

\State \Return $G=(V,E)$

\end{algorithmic}
\end{algorithm}

\begin{algorithm}[t]
\caption{\small Query-Conditioned Table Search}
\label{alg:search}
\scriptsize
\begin{algorithmic}[1]
\Require Query $q$, embedder, offline index, node embeddings, edge embeddings, join graph, learned scorer $S$
\Require Parameters: top-$K$, number of anchors $A$, max hops $H$
\Ensure Ranked list of tables

\State $x \gets \mathrm{Embed}(q)$
\State $C \gets \mathrm{FAISS\_Search}(x, K)$ \Comment{candidate nodes}
\State $A_0 \gets \mathrm{DeduplicateByTable}(C)$ \Comment{keep best node per table}
\State $D_E \gets \emptyset$ \Comment{edge deposits}

\ForAll{anchor table $T_a \in A_0$}
    \State $L \gets \{(T_a, s_a)\}$ \Comment{current frontier: table $\rightarrow$ score}
    \State $V \gets \emptyset$ \Comment{visited tables}

    \For{$h=1$ to $H$}
        \State $L' \gets \emptyset$
        \ForAll{$(T_f, s_f) \in L$}
            \State $V \gets V \cup \{T_f\}$
            \State $E_f \gets \mathrm{OutgoingJoinEdges}(T_f)$
            \If{$E_f = \emptyset$}
                \State \textbf{continue}
            \EndIf

            \ForAll{$e=(k_f,k_n)\in E_f$}
                \State Construct feature vector:
                \[
                \begin{aligned}
                \phi(e) = \Big(
                &x \,;\,
                \mathrm{emb}(e) \,;\,
                \text{MaxSim}(T_n), \\
                &\text{MaxSim}(T_f) \,;\,
                s_f \,;\,
                h
                \Big)
                \end{aligned}
                \]
            \EndFor
            \State $r \gets S(\phi)$
            \State $p \gets \mathrm{softmax}(r)$

            \ForAll{$(e,T_n,p_e)\in (E_f,\mathrm{nbrTables}(E_f),p)$}
                \State $d \gets s_f \cdot \alpha_h \cdot p_e$
                \State $D_E[e] \gets D_E[e] + d$
                \If{$T_n \notin V$}
                    \State $L'[T_n] \gets L'[T_n] + d$
                \EndIf
            \EndFor
        \EndFor
        \State $L \gets \{(T,s)\in L' \mid T \notin V\}$
        \If{$L=\emptyset$}
            \State \textbf{break}
        \EndIf
    \EndFor
\EndFor

\State $D_T \gets \mathrm{AggregateTableDeposits}(D_E)$
\State \Return $\mathrm{RankTables}(D_T)$
\end{algorithmic}
\end{algorithm}

\section{Hyperparameters}
\label{sec:appendix-hp}
Table \ref{tab:hp} lists the hyperparameters used across all main experiments, including values for anchor candidates, number of anchors, BFS hops, MMR weight, model dimensions, optimization settings, learning rate, weight decay, gradient clipping, learning rate scheduler, margin, training epochs, and cross-validation folds.
\begin{table}[!ht]
\centering
\small
\begin{tabular}{ll}
\toprule
\textbf{Hyperparameter} & \textbf{Value} \\
\midrule
Max BFS hops $H$                  & $5$    \\
Scorer hidden dim                 & $64$   \\
Dropout                           & $0.3$  \\
Optimizer                         & Adam   \\
Learning rate                     & $10^{-3}$ \\
Weight decay                      & $10^{-4}$ \\
Gradient clipping                 & $1.0$ \\
LR scheduler                      & \texttt{ReduceLROnPlateau} \\
\quad factor / patience           & $0.5$ / $3$ \\
Margin $m$                        & $0.1$ \\
Training epochs (BIRD)            & $20$ \\
CV folds (BEAVER)                 & $5$ \\
\bottomrule
\end{tabular}
\caption{Hyperparameters used in all main experiments.}
\label{tab:hp}
\end{table}

\section{Full Results on Spider}
\label{sec:appendix-spider}
Table~\ref{tab:spider_results} reports the complete Spider retrieval results omitted from the main paper for space considerations.

\begin{table}[t]
\centering
\scriptsize
\setlength{\tabcolsep}{4pt}
\begin{tabular}{l|cc|cc|cc}
\toprule
\multirow{2}{*}{Method} &
\multicolumn{2}{c|}{P} &
\multicolumn{2}{c|}{R} &
\multicolumn{2}{c}{PR} \\
& @5 & @15 & @5 & @15 & @5 & @15 \\
\midrule

\multicolumn{7}{c}{\textbf{BGE-large-en-v1.5}} \\
\midrule

Vanilla Retriever
& 0.215 & 0.072
& \textbf{0.745} & \textbf{0.752}
& 0.729 & \textbf{0.741} \\

REaR
& \textbf{0.216} & \textbf{0.099}
& \textbf{0.745} & 0.746
& \textbf{0.733} & 0.735 \\

\joingr{}$_{\text{Heuristic}}$
& 0.209 & 0.072
& 0.715 & 0.747
& 0.699 & 0.736 \\

\textbf{\joingr{}}
& 0.213 & 0.072
& 0.737 & 0.747
& 0.722 & 0.736 \\

\joingr{}$_{\text{BEAVER}\rightarrow \text{SPIDER}}$
& 0.210 & 0.072
& 0.721 & 0.747
& 0.705 & 0.736 \\

\midrule
\multicolumn{7}{c}{\textbf{all-MiniLM-L6-v2}} \\
\midrule

Vanilla Retriever
& \textbf{0.216} & \textbf{0.072}
& \textbf{0.749} & \textbf{0.752}
& \textbf{0.735} & \textbf{0.740} \\

REaR
& \textbf{0.216} & \textbf{0.121}
& 0.748 & 0.750
& 0.734 & 0.737 \\

\joingr{}$_{\text{Heuristic}}$
& 0.214 & 0.072
& 0.739 & 0.748
& 0.726 & 0.737 \\

\textbf{\joingr{}}
& \textbf{0.216} & 0.072
& 0.747 & 0.748
& \textbf{0.735} & 0.737 \\

\joingr{}$_{\text{BEAVER}\rightarrow \text{SPIDER}}$
& 0.214 & 0.072
& 0.739 & 0.748
& 0.725 & 0.737 \\

\midrule
\multicolumn{7}{c}{\textbf{Additional Baselines}} \\
\midrule

DTR
& 0.284 & 0.101
& 0.958 & \textbf{0.998}
& 0.930 & \textbf{0.994} \\

Contriever
& 0.417 & 0.143
& 0.970 & 0.994
& 0.928 & 0.987 \\

JAR$_{\text{F (DTR)}}$
& \textbf{0.419} & --
& \textbf{0.978} & --
& -- & -- \\

JAR$_{\text{F (Contriever)}}$
& 0.399 & --
& 0.937 & --
& -- & -- \\

IterativeJAR$_{\text{Contriever}}$
& 0.293 & 0.099
& 0.975 & 0.985
& \textbf{0.959} & 0.975 \\

\bottomrule
\end{tabular}

\caption{
Table retrieval results on \textbf{Spider}. 
}
\label{tab:spider_results}
\end{table}

\section{Reproducibility}
\label{sec:appendix-repro}

All training and evaluation use a single NVIDIA H200 GPU (40~GB), completing in under one hour for BIRD and under five minutes per fold for BEAVER. Frozen encoders are from Hugging Face Hub checkpoints \texttt{BAAI/bge-large-en-v1.5} and \texttt{sentence-transformers/all-MiniLM-L6-v2}. Full BFS-with-scorer evaluation runs at about $40$ queries/second on CPU once embeddings are cached, making \joingr suitable for online LLM-front-end use.

\end{document}